\documentclass[a4paper,11pt]{article}
\pdfoutput=1

\usepackage{jinstpub}
\usepackage{caption}
\usepackage{marvosym}

\usepackage{subfig}

\begin{document}
\newcommand{\ptt}{$p_T$}

\title{Adversarial domain adaptation to reduce sample bias of a high energy physics event classifier}

\author[*,1]{J.M.~Clavijo,}
\author[*,1]{P.~Glaysher}
\author[\text{\Yinyang},2, 3]{J.~Jitsev}
\author[*,\text{\Yinyang},1]{and J.M.~Katzy}
\note[*]{Equal contribution. {$^\text{\Yinyang}$} \text{Equal advising.}}

\note[*]{All figures and pictures by the author(s) under a CC BY 4.0 license.}

\affiliation{Deutsches Elektronen-Synchrotron DESY\\ Notkestra{\ss}e 85, 22607 Hamburg, Germany}

\affiliation[2]{Juelich Supercomputing Center (JSC), Institute for Advanced Simulation (IAS), Research Center Juelich (FZJ), Wilhelm-Johnen-Str., 52425 Juelich}

\affiliation[3]{Helmholtz AI, Research Center Juelich (FZJ), 52425 Juelich}

\emailAdd{judith.katzy@desy.de}

\abstract{We apply adversarial domain adaptation in unsupervised setting to reduce sample bias in a supervised high energy physics events classifier training. We make use of a neural network containing event and domain classifier with a gradient reversal layer to simultaneously enable signal versus background events classification on the one hand, while on the other hand minimising the difference in response of the network to background samples originating from different MC models via adversarial domain classification loss. We show the successful bias removal on the example of simulated events at the LHC with $t\bar{t}H$ signal  versus  $t\bar{t}b\bar{b}$ background classification and discuss implications and limitations of the method}


\maketitle

\section{Introduction}
\label{sec:intro}
Many measurements and searches for new phenomena performed by the experiments at the Large Hadron Collider (LHC) use a classification algorithm, such as Boosted Decision Trees or Neural Networks, to discriminate the physics process of interest (signal) from other physics processes with similar signature (background). The algorithms are optimized using supervised training on detailed Monte Carlo (MC) simulation data sets, containing samples labeled as signal or background. The resulting classifier is applied to unlabeled data to separate signal and background, and to measure the statistical significance of the signal or its strength, assuming that the simulated and the real data sets are identically distributed.

However, significant differences between domains of real and simulated data sets always exist and the learner may pick up those domain-specific discriminating features that perform well on classification task in one domain while being not suitable for classification in the other, introducing a bias via the source samples used for training when attempting to classify samples from target domain.  This problem is similar to that of visual recognition where, for instance, training may be performed on artificially generated images, the source domain, and applied to real photographs, the target domain. In order to avoid training a model that is suitable for classification on the source domain only, while failing when employed on target domain, algorithms of domain adaptation have been developed.

In this paper we apply the method of domain adaptation to a problem of classification on high energy physics data using a
Domain Adversarial Neural Network (DANN) ~\cite{ganin} to classify events in the search for the $t\bar{t}H(H\rightarrow b\bar{b})$ process at the LHC, which is very rare and hard to separate from the abundant $t\bar{t} +\text{jets}$
 background~\cite{ttHbb}. In the cited measurement work, a classifier is trained on labeled MC predictions to separate signal from background. The trained classifier is applied on MC where signal and background events are mixed according to the theoretical predicted fraction, and on data to obtain binned distributions of classifier output. The ratio of the resulting spectra is used in a profile likelihood fit to measure the signal ratio in data.
 The effect on the final result caused by the bias for the specific MC background model of the source domain used for training is estimated using an alternative simulation of a target domain, based on a different physics model, which was not used for training. The difference between the classifier outputs of the different  background MCs is taken as uncertainty on the classification in the fit. This uncertainty happens to be the largest on the measurement, hampering the observation of the searched process. Therefore a solution to minimize this sample bias is of high importance. For the study presented here, the two background simulations correspond to the different domains. The domain adaptation is applied to reduce mentioned training bias.

The network structure consists of a common feature extractor part and separate branches for label classification and domain adaptation, implemented via a gradient reversal layer as presented in~\cite{ganin}. This network structure differs from other adversarial approaches by including domain adaptation in the learning process via the shared feature extractor part used by both label and domain classifier as proposed on theoretical grounds in~\cite{bendavid}.  This way, the network model is pushed to extract discriminant features for the classification that are at the same time invariant to the different domains. The use case presented here differs from~\cite{ganin} as we provide a set of physical jet properties instead of images as input and the use of a bigger and more complex data set. Additionally, we make detailed performance analyses, evaluating the influence of several hyper-parameters and also exploring training issues that appear for this kind of architecture.

Adversarial classifiers without domain adaptation were used in high energy physics before, e.g.  to reduce theoretical uncertainties~\cite{englert}, to  decorrelate  a jet tagger from the jet mass~\cite{JetTag} and to tune  a classifier against a nuisance parameter~\cite{Pivot}.  An adversarial set-up involving domain adaptation without labels has been used for multi-class classification  in a search for long lived particles~\cite{Sirunyan:2019nfw}. In this paper,  we systematically study the algorithm  taking advantage of the two domains being labelled to control the results achieved without use of labels.  We address the challenges for learning algorithms in domain adaptation, the dependence of the hyper-parameters specific to domain adaptation and potential bias of the classifier.  Furthermore, since the use of domain adaptation without labels for multi-modal distributions can be a problem as pointed out in e.g. ~\cite{conditional}, we chose to  use  the  domain adaptation algorithm  for binary classification and proof its applicability for this use case in contrast to the multi-class application mentioned above.

In this paper we describe the network used in Sec.~\ref{sec:architecture}, followed by the details of the data sets used in Sec.~\ref{sec:datasets}.
We systematically study the dependence on hyper parameters in Sec.~\ref{sec:setup}, including some issues observed during the training. In Sec.~\ref{sec:results}, we expose the performance through different figure of merits related to physics searches and we include a feasibility study for a potential use with real unlabeled data. Finally, a summary and some conclusions are given in Sec.~\ref{sec:conclusion}.

\section{The Deep Adversarial Neural Network}
\label{sec:architecture}

\begin{figure}[h]
	\includegraphics[width=1.0\textwidth]{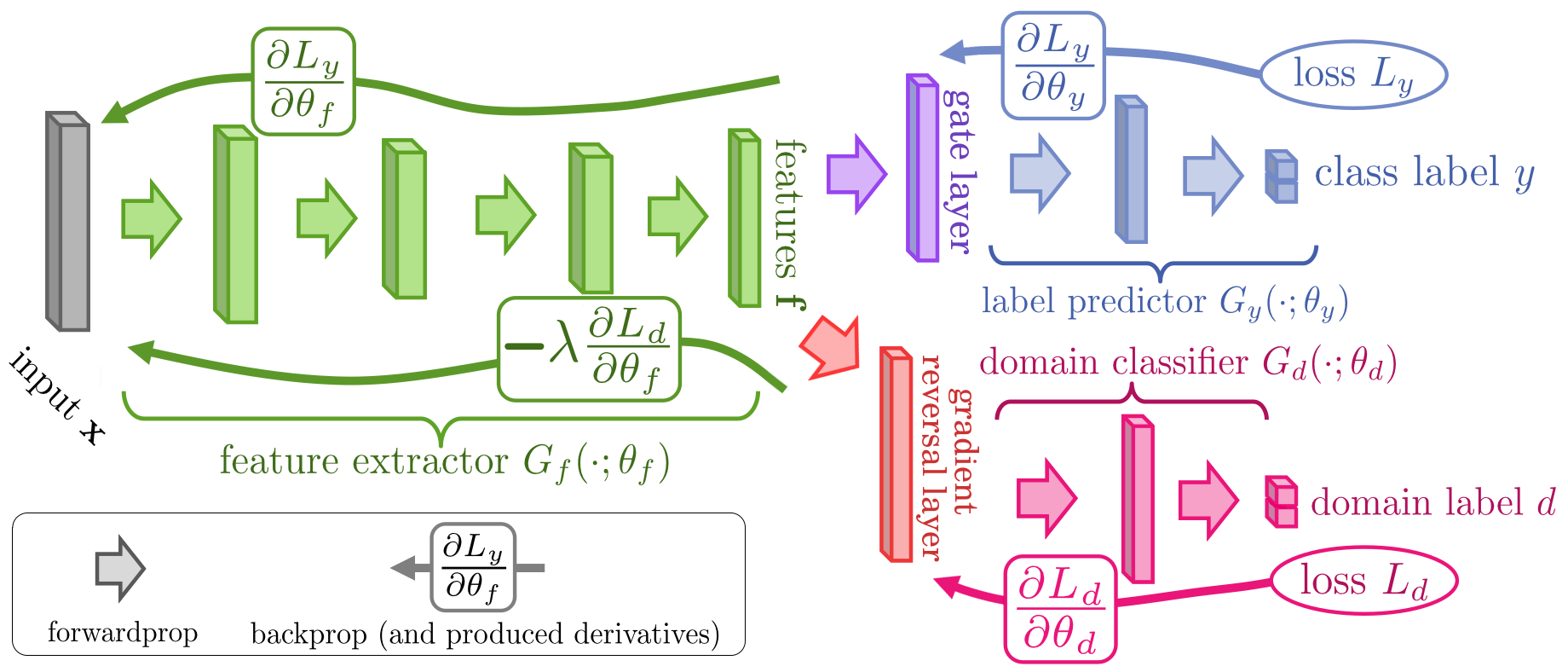}
	\caption{Domain-adversarial network as an alternative to reduce classification bias, adapted from~\cite{ganin}.}
	\label{fig:Network}
\end{figure}

We follow the architecture presented in \cite{ganin} with a feed-forward neural network composed of three parts as shown in Fig.~\ref{fig:Network}: a \emph{feature extractor} which splits into the \emph{label predictor}, performing the signal-background classification, and the \emph{domain classifier}. Domain adaptation is enabled via an adversarial interplay between domain classifier and feature extractor. For training and testing we have two data sets (domains): source and target, both containing signal and background events. The target domain is constructed as a representative pseudo-data, meaning that it is treated as unlabeled and it has a signal to background proportion similar to the one expected in a real data sample. For measuring our algorithm performance we make use of the target labels in the final test.

For the label classification we train the network only using events from the source domain. The gate layer stops the target events propagation making the \emph{label predictor} loss being evaluated only on the source events. This allows training the network on mixed samples of both domains. The classification is adapted to the target domain by connecting the \emph{feature extractor} with the \emph{domain classifier} through a gradient reversal layer. This layer does nothing during the forward propagation but inverts the sign of the gradients flowing from domain classifier during the backpropagation. The \emph{domain classifier} is trained to determine which domain the events belong to. Due to the gradient reversal, the \emph{feature extractor} is pushed to form such feature representation that do not allow to distinguish between two different domains, thus avoiding the sample bias.
As a result of such adversarial training, the features in the last layer of the \emph{feature extractor} will both allow the classification between signal and background and become domain invariant, rendering classification model domain-independent. The gradients of the reversal layer are scaled by the parameter $\lambda$ allowing to regularize their influence and hence tune the importance of the label classification versus the domain invariance.

In order to have balanced classes for each classification the event weights of the source domain are scaled as required according to class ratios. For the label predictor the weights are such that the effective number of signal and background events are the same. For the domain classifier, the weights are scaled to match the signal to background ratio existing in the target domain.

\section{Data sets}
\label{sec:datasets}
The feature selection for the input of the network was inspired by the analysis presented in \cite{ttHbb} to separate $t\bar{t}H$ from the $t\bar{t}+b\bar{b}$ background. In total 41 geometrical and kinematic quantities are used as input to the network, such as the angular distance between different jets and/or leptons, the mass of various jet and lepton systems and the event topology. The complete list of features, their correlations and the relative importance are given in ~\cite{Ranking}.

We use MC samples provided by the HepSim Group~\cite{opendata}. The ttH signal sample containing $13\cdot 10^6$ events was generated with MadGraph~\cite{Alwall:2011uj} matched to the Herwig6 parton shower~\cite{Corcella_2001}. Two background samples were generated, significantly differing in the theoretical predictions. One, used for the source domain, with $2\cdot 10^6$ events of top quark pair production with additional light quarks using MadGraph matched to the Herwig6 merged with $10^7$ top quark pair events with additional bottom-quarks using MadGraph matched to Pythia6~\cite{Sjostrand:2006za}. The other background sample, which is used for the target domain, contains $3\cdot 10^7$ events of top quark pair production in association with bottom quark pairs, generated with the PowhegBox+OpenLoops~\cite{Tomas} and is matched to Pythia8 for the full event generation including the prediction of additional light quarks.

The ATLAS detector response  was simulated using Delphes simulation~\cite{delphes}.  For this study, reconstructed leptons, jets and bottom\footnote{$bottom$ stands for bottom and anti-bottom quarks} quark iniated  jets (called $b$-jets in the following) are used. Jets are reconstructed using the anti-kT algorithm \cite{antikt} with a radius of $R=0.4$. The identification efficiency of $b$-jets was taken from~\cite{Aaboud:2018xwy}, assuming the reconstructed $b$-jets to have a 70\% tagging probability with a corresponding light jet/c-jet rejection probability parameterization.

Events selected for the neural network training were required to fulfil the following criteria:
\begin{itemize}
	\item one electron or muon with transverse momentum \ptt $\geq$ 20\,GeV
	\item at least 5 jets with \ptt $\geq$ 25\,GeV
	\item at least 3 $b$-jets.
\end{itemize}

With this selection applied the source and target data sets where constructed with $546\cdot10^3$ signal each and same amount of background events, using statistically independent events from the same simulation as signal but different background simulations for source and target. One half from each data set was left for testing purposes, the remaining were used for training. For the target domain only $14368$ signal events were randomly selected for training, to match the 5:95 ratio of signal to background estimated in real data.

\section{Network set-up and training}
\label{sec:setup}

The network was implemented using the Keras v2.2.4~\cite{keras} with TensorFlow v1.12~\cite{tensorflow} as back-end library. The training set-up is described in Sec.~\ref{sec:general-setup}. A hyper-parameter scan was done to optimize the performance of the network, as described in Sec.~\ref{sec:hyperparams}. Some special considerations for the loss function and its optimization are described in  Sec.~\ref{sec:loss_function} and  Sec.~\ref{sec:optimiser}, respectively.

\subsection{Training set-up}
\label{sec:general-setup}
The initial weights of the network were set by the Xavier initializer, as suggested in~\cite{xavier}. The number of training epochs was dynamically selected with the following condition applied: the training were stopped if the running average over 50 epochs in the total loss does not decrease more than 0.05\% with respect to the previous 50 epochs. This number was restricted to the interval [200, 1000]. The lower limit was set to skip some random fluctuations at the beginning. The upper limit is just a big value that was never reached with the specified condition. After the training was stopped the weights of the network in the epoch with the lowest label predictor loss were selected. A batch size of 16384 was used. Each batch was composed by source and target events in a 1:1 proportion. The events were randomly shuffled at each epoch, resulting in a different batch selection each time. The \emph{domain classifier} and \emph{label predictor} outputs were set to have two neurons each, using softmax activation function and cross-entropy loss in both (Sec.~\ref{sec:loss_function} describes an alternative). The \mbox{RMSProp} Keras optimizer was used, with the parameters: learning\_rate =0.001 and rho=0.9.

\subsection{Hyper-parameter optimization}
\label{sec:hyperparams}

The hyper-parameters of the network were chosen with the help of the Hyperopt library~\cite{hyperopt}, using the Tree of Parzen Estimators (TPE) algorithm implemented on it. The number of layers in each part of the network was let vary from 1 to 8. Each layer could have a number of neurons between 5 and 100, but having a linear behavior in each part of the network (either decreasing or increasing). For the activation function of the hidden layers ReLU, tanh and ELU were tested. Each of this hyper-parameters were sampled from a uniform distribution. Additionally, the $\lambda$ parameter was sampled from a log-uniform distribution in the range [1, 1000], with this giving more priority to low values as these were found to give better results.

The additive inverse of the label \emph{label predictor} area under the receiver operating characteristic curve for the target domain was used as the loss to minimize. Three independent optimizations where performed in parallel in order to have a better view of the hyper-parameter space.
This also helps to detect if the global minimum of the loss is found.
Approximately 1000 iterations where performed in each case.

By analyzing the sets of parameters with good performance and the decisions made by the sampling algorithm, we were able to draw the following conclusions:
\begin{itemize}
	\item The optimal number of layers in the \emph{label predictor} is one: only the output layer. Two is also good in cases of a very complex \emph{feature extractor}.
	\item Higher complexity in the \emph{feature extractor} provides performance improvement but also makes the network more prone to over-training.
	\item The number of neurons in the last layer of the feature extractor should be at most the same that in the input. We think this number is also related to the correlations in the input features: a smaller number for high correlations could provide a better optimized feature extraction.
	\item An increase in the domain classifier complexity does not cause significant improvements, but it needs at least a similar complexity than the feature extractor in order to provide good corrections.
	\item The performance with ELU and tanh as activation function for the hidden layers was very similar. ReLU was significantly worse.
\end{itemize}

Finally the \emph{feature extractor} was chosen to have four layers with 20, 16, 13 and 10 neurons respectively, the \emph{label predictor} with only the output layer (2 neurons) and the \emph{domain classifier} with four layers of 20, 35, 50 and 2 neurons respectively. The ELU activation function was used in all the hidden layers.

Note that due to the non-deterministic nature of the training process, results during the optimization were sometimes not representative of the behavior for each set of hyper-parameters tested. Set-ups with higher performance were found, but its results were not reproduced in further tests. Therefore, we chose a configuration with stable results instead of the best one reported by the optimization process. It also had the advantage of being not complex enough to be affected by over-training.

\subsection{Loss and activation functions for the outputs}
\label{sec:loss_function}

The total loss of the network ($L$) is given by the sum of the individual losses of the \emph{label predictor} ($L_y$) and \emph{domain classifier} ($L_d$):
\begin{equation}
L = L_y  + L_d
\end{equation}
The gradient reversal layer affect the backpropagation in such a way that the gradients of the total loss with respect to the \emph{feature extractor} weights ($\theta_f$) are computed as:
\begin{equation}
\frac{\partial L} {\partial \theta_f} = \frac{\partial L_y} {\partial \theta_f} - \lambda \frac{\partial L_d}{\partial \theta_f}
\end{equation}

Two alternatives were used for computing  the loss: set-up~A with a softmax activation and cross-entropy loss in both outputs, and set-up~B with softmax and cross-entropy loss in the \emph{label predictor}, and  linear loss following eqn.\ref{eqn:linearloss} in the \emph{domain classifier}.

The cross-entropy loss for a single event $E_i$ is given by:
\begin{equation}
L_i =
\begin{cases}
-\ln(y_i)    & \text{if } E_i \in \text{class 1}  \\
-\ln(1-y_i) &  \text{if } E_i \in \text{class 0}
\end{cases}
\label{eqn:crossentropy}
\end{equation}
where $y_i$ represents the network output for that event. Note that even though we have a two-neuron output we refer to $y_i$ as a single value since the second neuron behaves as 1 minus the first. Class 0 corresponds to background and class 1 to signal for the \emph{label predictor}. A perfect classification yields a loss of 0, value toward which the loss is optimized.

Set-up~A also uses this loss for the \emph{domain classifier}, with $y_i$ in equation~\ref{eqn:crossentropy} corresponding to the \emph{domain classifier} output and classes~0 and 1 corresponding to target and source domains respectively. In this case, perfect separation also results in a loss of 0 but a separation between the domains is not intended. Instead, the network response should be the same for both classes of events which is provided as an additional restriction via the gradient reversal layer. The \emph{domain classifier} loss is minimized but, under this restriction, the lowest achievable loss is when the response for both classes, i.e. source and target, is $y_i=0.5$, resulting in a loss of ${-\ln{0.5}\approx 0.693}$. This behavior is visible in Fig.~\ref{fig:fluctuations}b, where the predicted loss of 0.693 is reached in the first epochs and kept most of the training. It should be noted that this poses an extra requirement on the \emph{feature extractor}, which besides providing domain independent features, is also optimised to provide features for which the \emph{domain classifier} output are exactly 0.5 for all events.

We found that deviations in the output of the \emph{domain classifier} from the optimal value of $y_i=0.5$ had severe influences on the classification in general. Analyzing at a lower level we found that these changes were driven by huge gradients back-propagated from the domain classifier loss, further amplified by $\lambda$ as $\lambda > 1$ was used. To avoid the change in the gradients under $y_i$ deviations we tested a set-up where the derivatives of the \emph{domain classifier} loss were independent of the $y_i$ values. To achieve this behaviour, we removed the activation function from the \emph{domain classifier} output and changed the loss to a linear function, computed for a single event ($E_i$) as:

\begin{equation}
	L_i =
	\begin{cases}
		-y_i  & \text{if } E_i \in \text{source}  \\
		\ \; \, y_i & \text{if } E_i \in \text{target}
	\end{cases}
	\label{eqn:linearloss}
\end{equation}
This new set-up has also the advantage that $y_i$ is not limited to 0.5 in the optimized case, since now, if the condition of no domain separation is met, this loss has a  value of 0 for any value of the \emph{domain classifier} output so the \emph{feature extractor} has more freedom during the optimization.

\subsection{Training of the neural network }
\label{sec:optimiser}

\begin{figure}[h]
	 \subfloat[]{\includegraphics[width=0.5\textwidth]{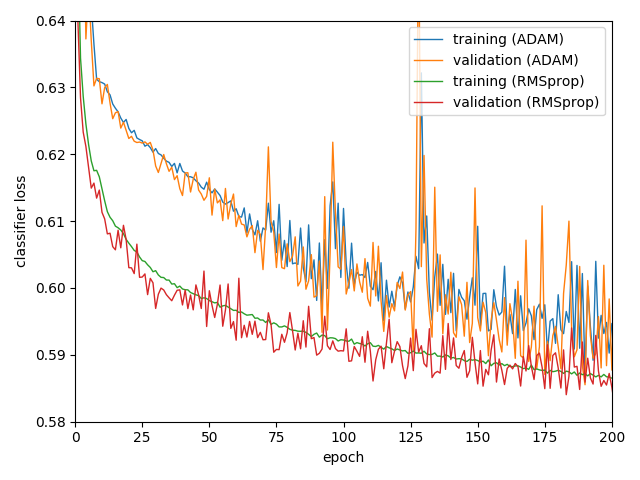}}
	  \subfloat[]{\includegraphics[width=0.495\textwidth]{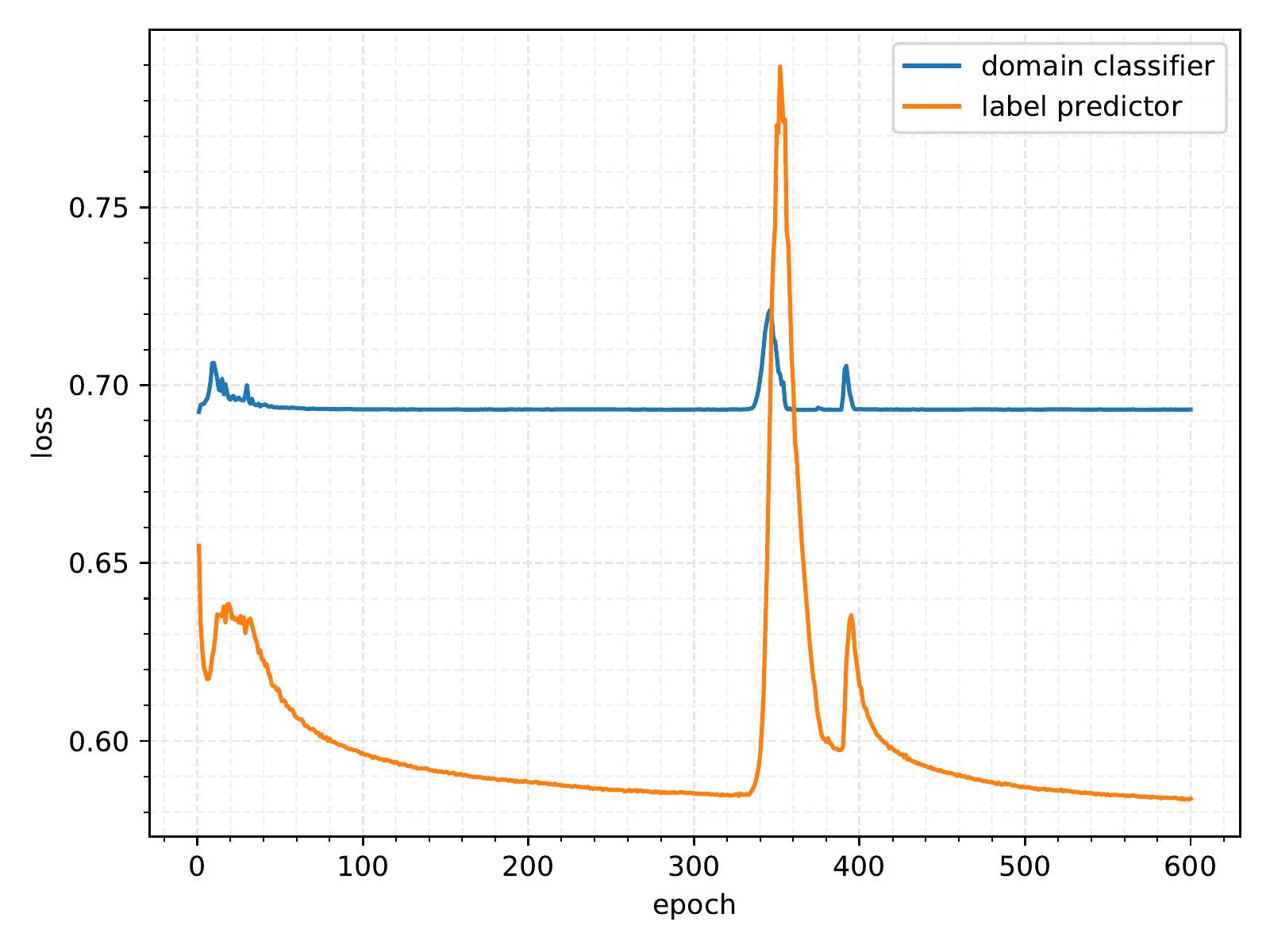}}	\caption{Examples of fluctuations observed during training of the network using set-up~A. (a) Training and  validation loss for the label predictor using the ADAM and RMSprop optimizers. (b) Big spikes of the label predictor and domain classifier losses (case with $\lambda=20$).}
	\label{fig:fluctuations}
\end{figure}

The ADAM optimizer \cite{adam}, being commonly used nowadays, was used as starting point. ADAM is an extension of RMSProp with SGD Momentum i.e. adding  momentum terms  defined as decaying average of the past gradients. The momentum terms should help to faster escape from highly sub-optimal loss regions.  However, when we used the default values of the momentum term ($\mu = 0.9$) we noticed severe oscillations of the label predictor loss, as shown in Fig.~\ref{fig:fluctuations}a. These oscillations seem to be caused by fluctuations in the domain classifier loss part on which the label predictor has then to react in the common effort to minimize the global loss. We  switched to RMSProp, which does not use a momentum term, resulting in a more stable loss course during the training.  We therefore did not attempt to further use ADAM.

Beside those small fluctuation described above, infrequent large spikes where found. One of them is shown in Fig.~\ref{fig:fluctuations}b, where $L_{y}$ minimizes smoothly for over 300 epochs but suddenly $L_{y}$ raise to huge values together with $L_{d}$.
Running 3000 independent trainings we found that these spikes appear in around a 0.7\% of the cases for set-up A and 1.6\% for set-up B.

Performing analyses we found that changing the weights of the network to the ones used 10 epochs before makes the spikes vanish. This indicates that the cause of the spikes involves initial random fluctuation related to the adversarial training with the gradient reversal layer. Backing up this assumption, we also found that the frequency of these spikes increases by increasing the value of $\lambda$.

\begin{figure}[h]
	\subfloat[]{\includegraphics[width=0.5\textwidth]{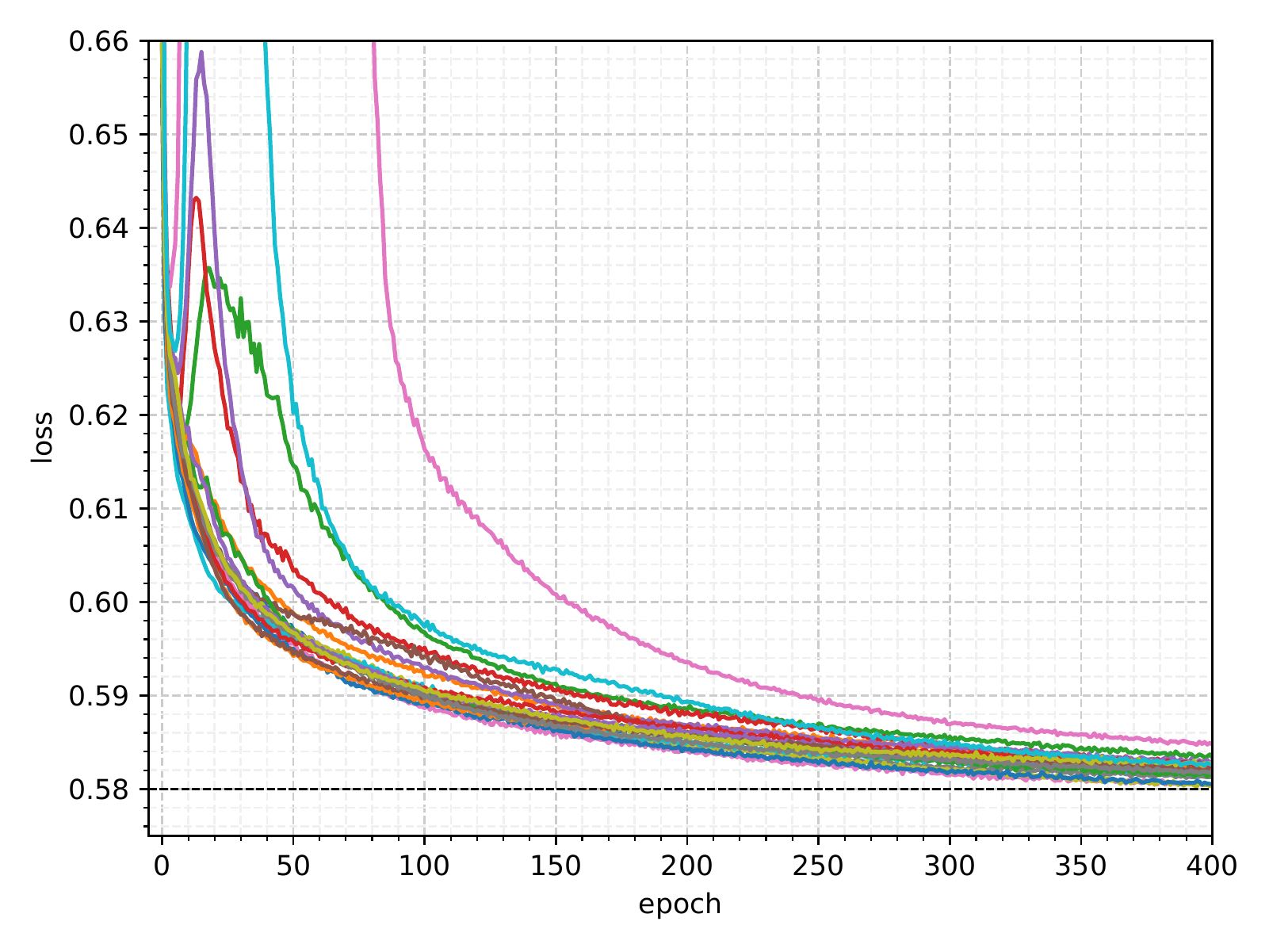} }
	\subfloat[]{\includegraphics[width=0.5\textwidth]{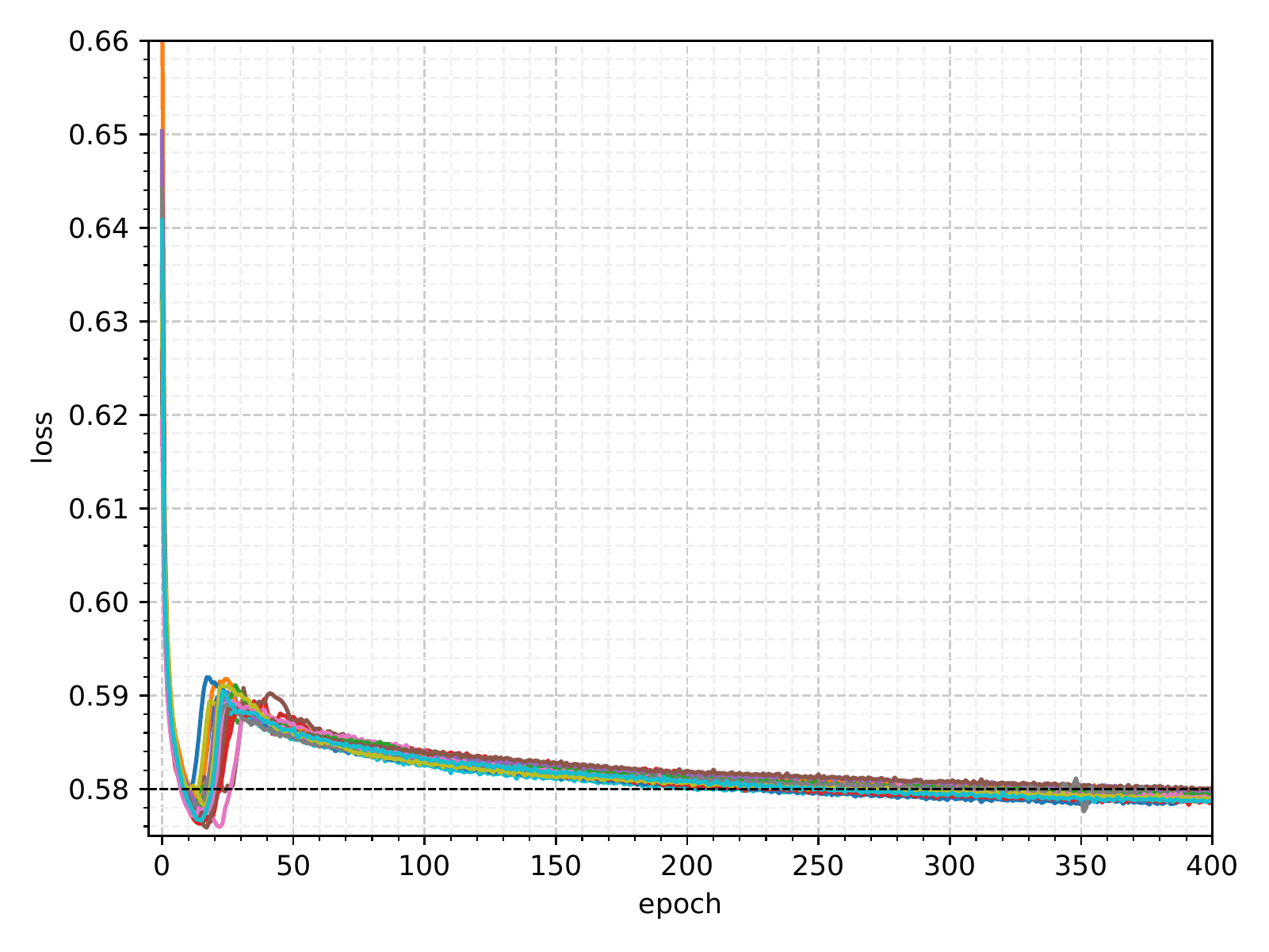} }
	\caption{\emph{Label predictor} loss in 20 different trainings starting from different random seeds for (a) set-up A with $\lambda = 20$ and (b) set-up B with $\lambda = 10^{-5}$. For better visibility, the y-axis is limited to 0.66 cutting off high fluctuations. Note that set-up B reaches better \emph{label predictor} loss.}
\label{fig:convergence}
\end{figure}

Furthermore, comparing set-ups A and B, we found in A stronger dependence of the learning curves on the randomization (initial weights, data shuffling, etc.), which is demonstrated in Fig.~\ref{fig:convergence}. The learning curves for set-up B agree better indicating a more stable training. They also converge faster.
The stopping criterion is reached in set-up A after around 600 epochs but after about 400 epochs in set-up B.

\subsection{Tuning impact of the adversarial domain classification}
\label{sec:lambda_scan}

\begin{figure}[h]
	\subfloat[]{\includegraphics[width=0.5\textwidth]{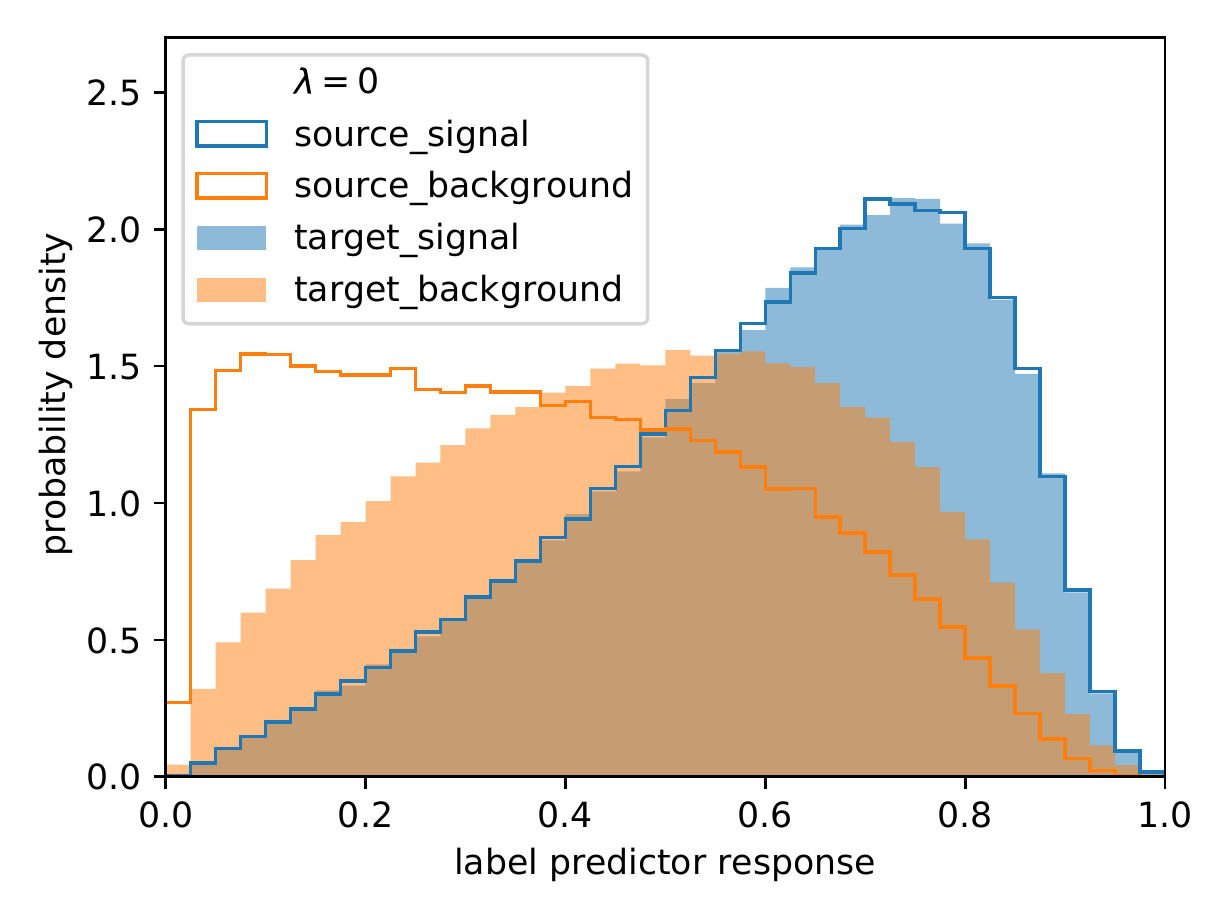} }
	\subfloat[]{\includegraphics[width=0.5\textwidth]{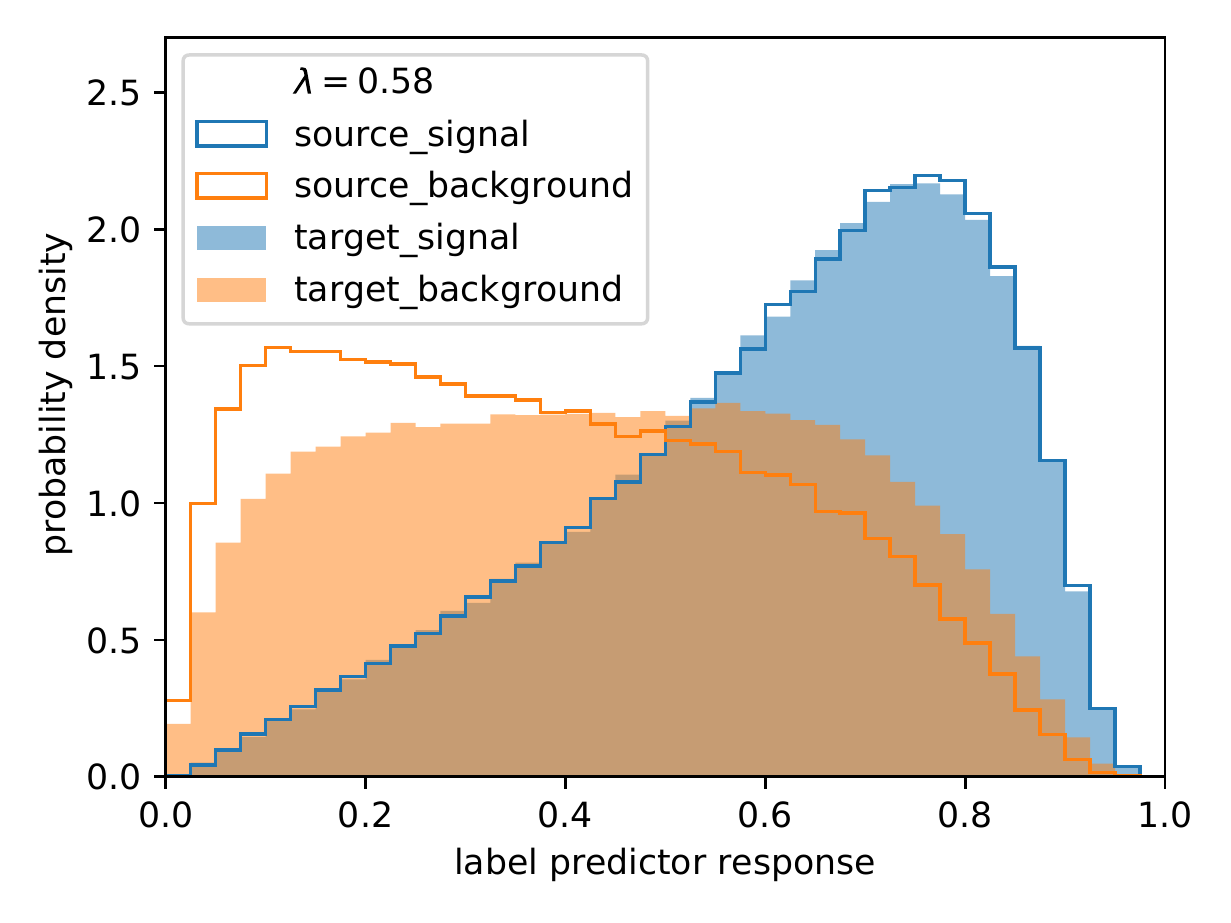} }
	\
	\subfloat[]{\includegraphics[width=0.5\textwidth]{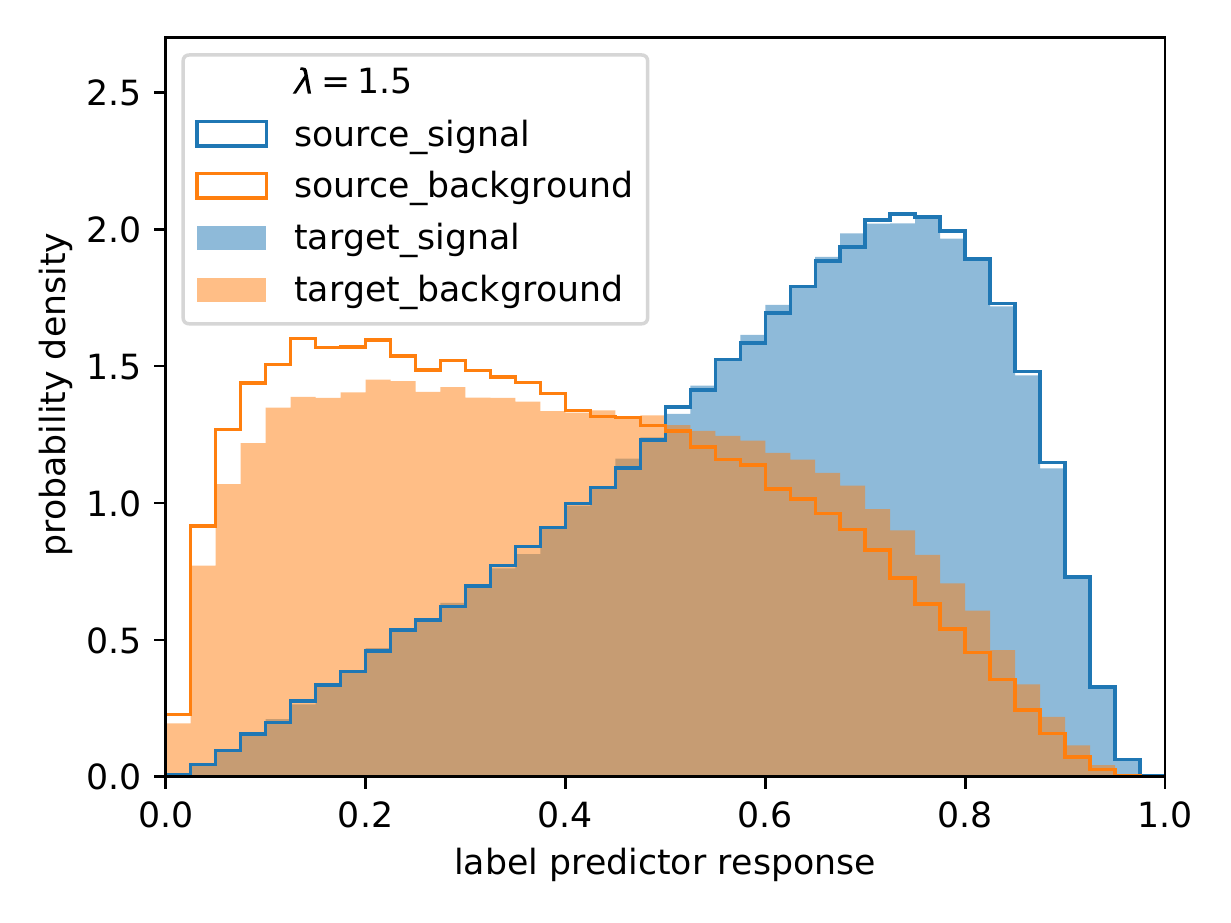} }
	\subfloat[]{\includegraphics[width=0.5\textwidth]{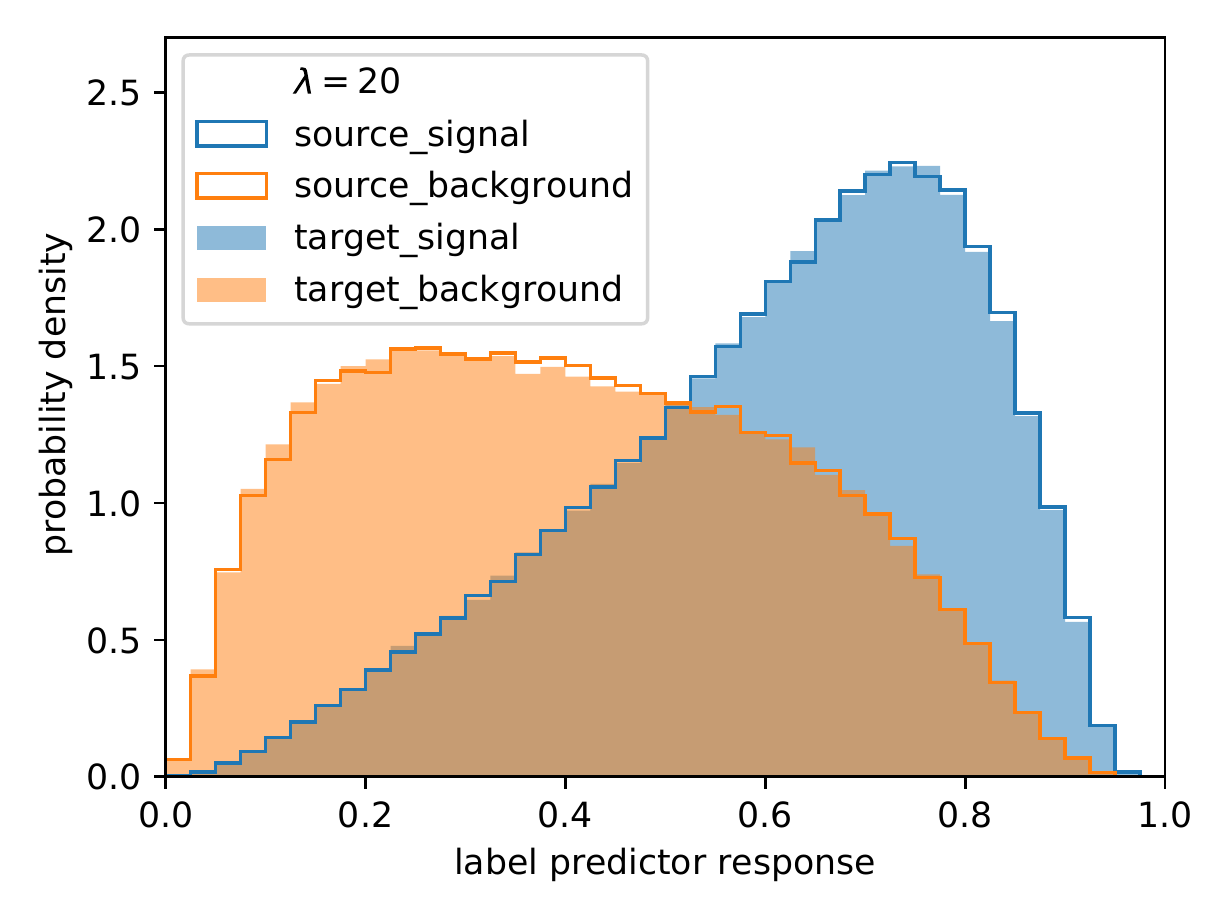} }
\caption{Label predictor response to signal (blue) and background (orange) for different values of   $\lambda$.  The label predictor is trained on the source domain and applied to a statistically independent part of the source domain (lines) and the target domain (area). Each of the distribution is normalized to 1. (a) $\lambda=0$    (b) $\lambda=0.58$    (c) $\lambda=1.5$
	( d) $\lambda=20$ . Discussion see text.}
	\label{fig:Response}
\end{figure}

The parameter $\lambda$ controls the influence of the label predictor and domain classifier responses on the total loss. It determines how much the responses to source and target data input produced by feature extractor should agree. High $\lambda$ values forces a strong agreement but may impair the ability of the feature extractor to provide useful features for the classification, low values give more freedom for the feature representation density distribution but might not be enough for obtaining a good agreement between the domains and thus introduce a bias for source domain samples. To give an example, Figure~\ref{fig:Response} shows the discriminant output for the set-up A for values of $\lambda$ between 0 and 20.  A large difference between source and target domain feature extractor response density can be observed for $\lambda=0$, while with increasing values of $\lambda$ the influence of the domain classifier on the density alignment and consequently also on label prediction increases and finally a very strong agreement between feature extractor responses to both background samples is reached at the highest value of $\lambda$, while label predictor performance deteriorates.
The optimal lambda value is specific to the problem and the performance measure applied as will be discussed in the following.

\begin{figure}[ht]
	\subfloat[]{\includegraphics[width=0.5\textwidth]{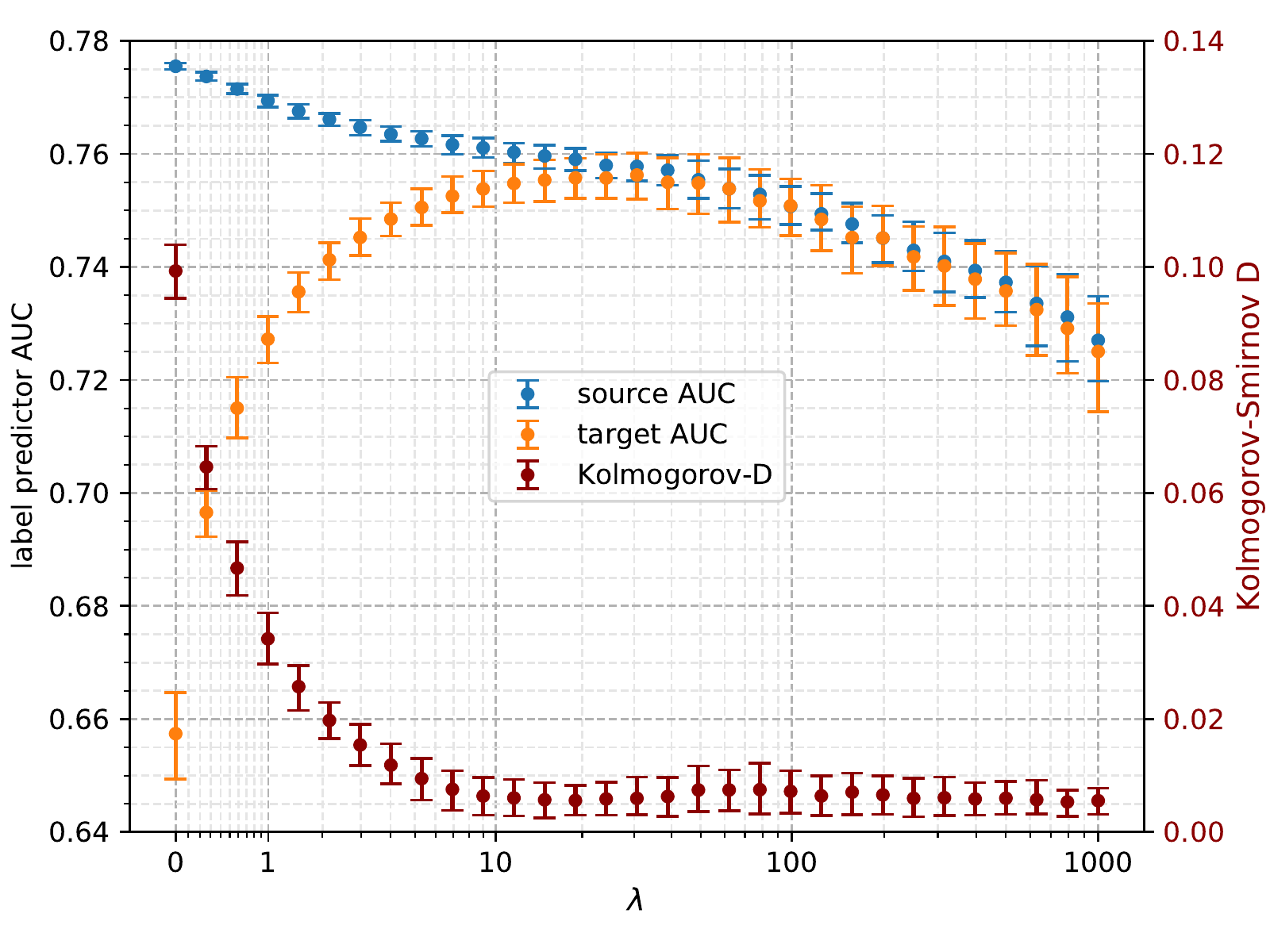} }
	\subfloat[]{\includegraphics[width=0.5\textwidth]{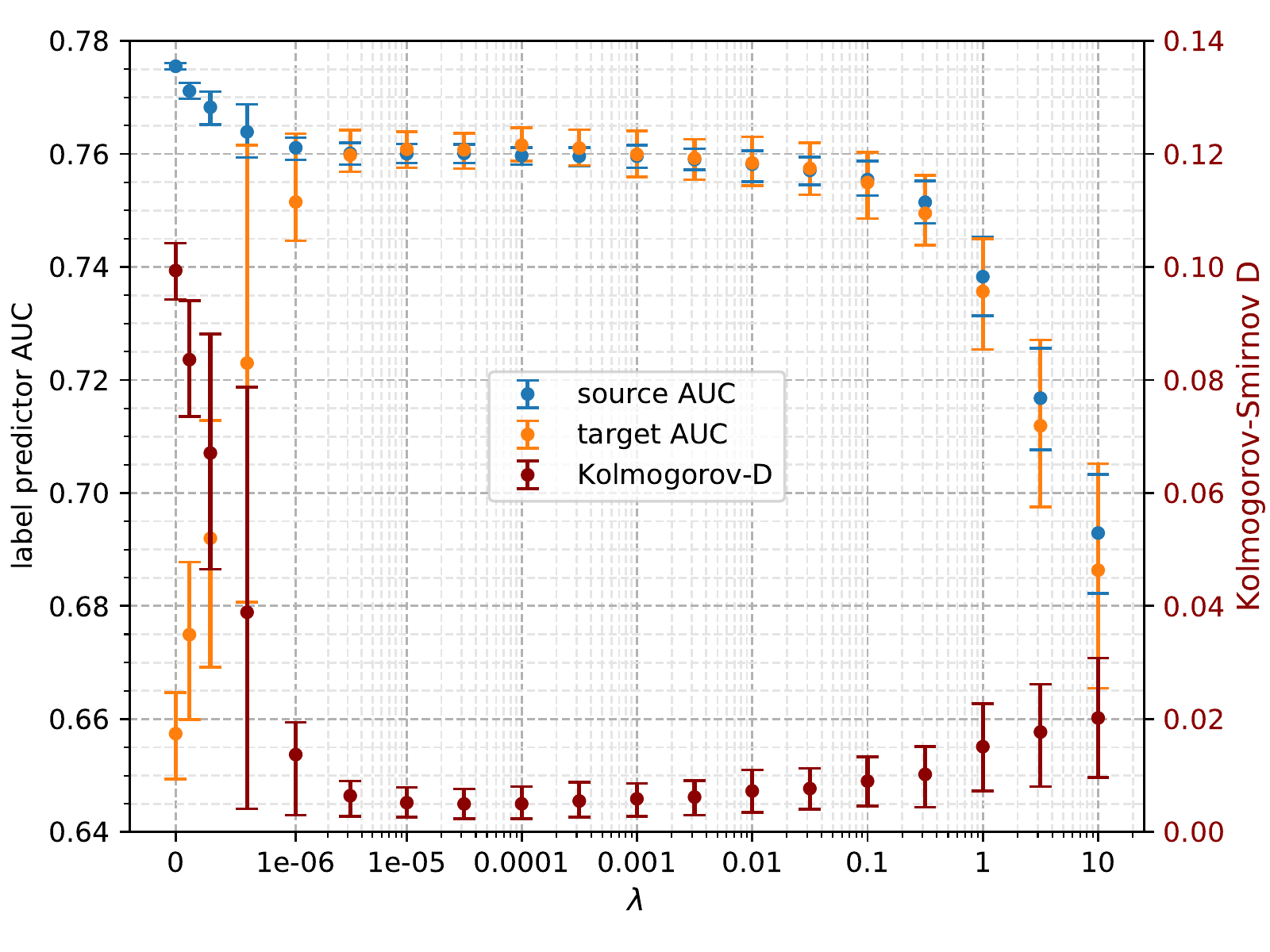}}
	\caption{Performance measured as the area under the ROC curve (AUC) for several values of $\lambda$. The difference between the response for source and target is measured as the Kolmogorov-Smirnov distance. (a) set-up~A  (b) set-up~B. Each point represent the average over $\sim 200$ independent training processes (with different random numbers). The error bars represent the 15.8 and 84.2 percentiles, corresponding to $\pm 1\sigma$ in a normal distributed variable.}
	\label{fig:lambda_scan}
\end{figure}

\section{Results}
\label{sec:results}

The performance of the network  depends on the relative importance of the adversarial branch containing domain classifier steered by the parameter $\lambda$. As for any hyper-parameter, the values of $\lambda$ are specific to the network architecture and data sets used and need to be determined for each particular use case.
We consider three measures of performance, demonstrating the bias without the adversarial treatment and their improvement when the adversarial branch is included.

First we report AUC which is a common performance measure for binary classifiers. Since a good value for $\lambda$ was still not selected we made a scan over a range of possible values (Fig.~\ref{fig:lambda_scan}). We extend it with the Kolmogorov-Smirnov distance as a measure of agreement between the response of the two domains. This distance is given by the maximum absolute difference between the cumulative distributions of the normalized \emph{label predictor} response for the two domains. The best choice of $\lambda$ is the value for which the maximum source domain AUC is achieved among those with the lowest Kolmogorov-Smirnov distance. This criterion for the optimal $\lambda$ has the advantage that it can be computed without using the target labels, i.e. using labeled source and unlabeled target data. To demonstrate that the criterion for $\lambda$ selection leads to desired behavior on target data, we compute the AUC for the target domain, as in our study target labels were provided by the simulation. As depicted in Fig.~\ref{fig:lambda_scan}, the closest match between source and target domain and highest AUC performance is achieved when using lambda values obtained from the criterion procedure.

With $\lambda = 0$, corresponding to absence of adversarial network, an AUC on the target domain of 0.657 is achieved.
This value is improved to 0.756 using $\lambda = 20$ for set-up A, and 0.760 using $\lambda = 10^{-5}$ for set-up B. This improvements have the cost of reducing the AUC obtained for the source domain from 0.776 in the no adversarial case, to 0.757 and 0.760 for set-ups A and B respectively with the selected $\lambda$ values. Increasing $\lambda$ above those values only decreases the performance, but in the case of set-up~B a plateau exist such that taking $\lambda$ values up to 100 times the selected one keeps the same performance.

\begin{figure}[h]
	\subfloat[]{\includegraphics[width=0.5\textwidth]{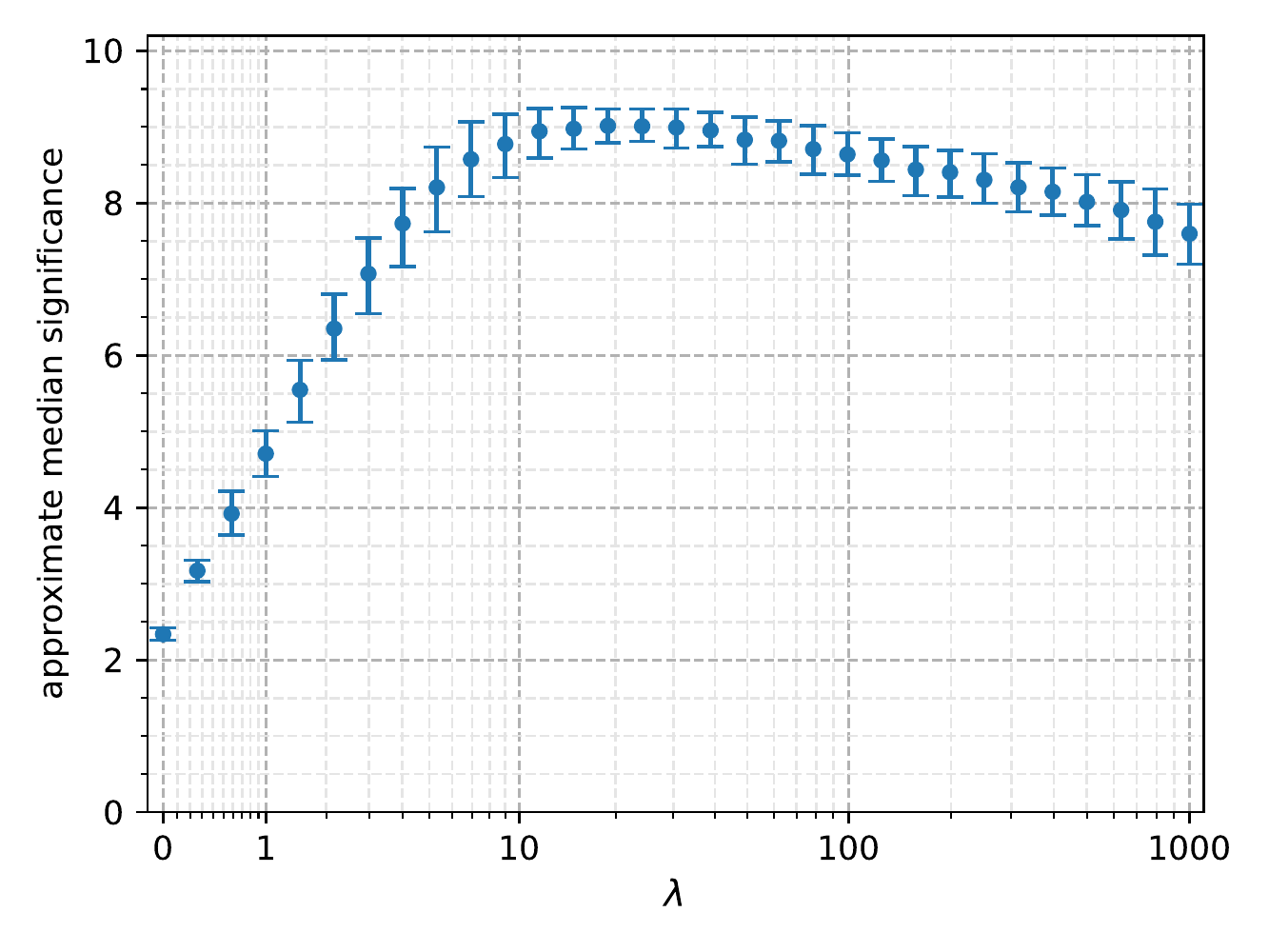}}
	\subfloat[]{\includegraphics[width=0.5\textwidth]{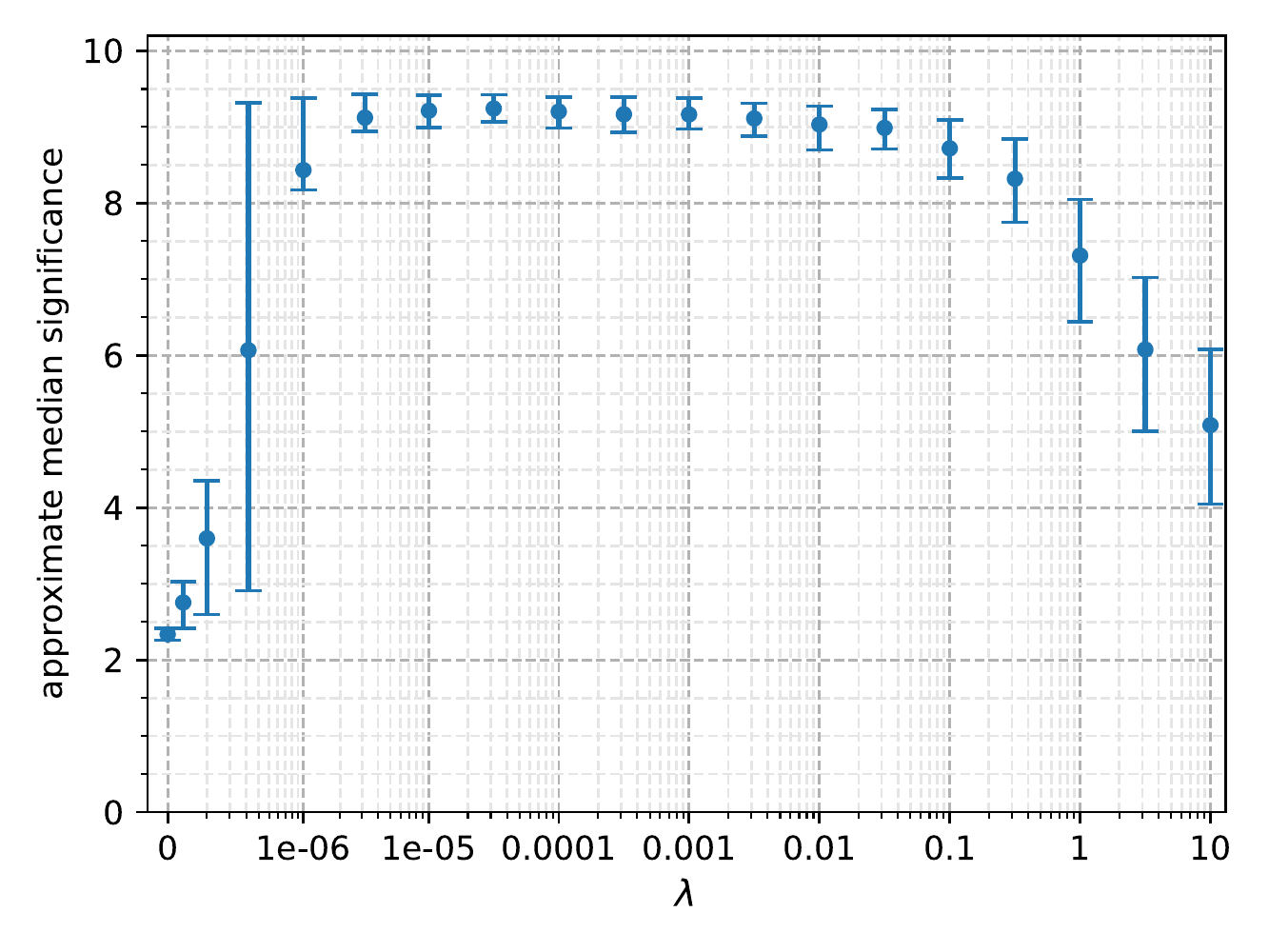}}
	\caption{Approximate median significance (in units of standard deviations) as a function of $\lambda$, computed for 50000 events consisting of 5\% signal and 95\% background. (a) set-up A (b) set-up B. Each point represent the average over $\sim 200$ independent trainings and the error bars represent the 15.8 and 84.2 percentiles.}
	\label{fig:significance}
\end{figure}

To further approximate the significance as reported in Higgs discovery searches as performance measure, we use the approximate median significance~(AMS) as proposed in~\cite{cowan}. This definition corresponds to a test of the signal discovery versus background only hypothesis by taking systematic uncertainties into account. It is calculated as:
\begin{subequations}
\begin{align}
\text{AMS} &= \sqrt{\sum_i \left\{ 2\left[ (s_i+b_i) \ln{s_i+b_i \over b_{0i}} - s_i - b_i + b_{0i} \right] + {(b_i - b_{0i})^2 \over \sigma_{bi}^2}\right\} } \\
b_{0i} &= {1 \over 2}\left( b_i - \sigma_{bi}^2 + \sqrt{(b_i - \sigma_{bi}^2)^2 + 4 (s_i + b_i) \sigma_{bi}^2} \right)
\end{align}
\end{subequations}
where the sum is over the bins in the histogram of the response, $s_i$ and $b_i$ represents the signal and background counts in the bin $i$ for the source domain and $\sigma_{b_i}^2= {1 \over 2}(b_i - b_i^{\ alt.})^2 + (0.1\ b_i)^2$ is an estimator of the variance on the background counts. The variance is computed from the difference between $b_i$ and the background count for the target domain in the same bin ($b_i^{\ alt.}$) plus a flat 10\% uncertainty on the background, approximating the values of the reference analysis. The AMS is a valid simplification of the significance in the context of this paper as long as we consider only the qualitative behavior, not the absolute values.  The AMS as a function of $\lambda$ is shown in Fig.~\ref{fig:significance}.  A low significance is observed in the case when the response for both domains disagree. The significance increases with $\lambda$  until reaching a maximum at similar positions of the maximal AUC where source and target values agree (Fig.~\ref{fig:lambda_scan}). For higher values of $\lambda$ the significance decreases, reflecting the loss of classification power.

\begin{figure}[h]
	\subfloat[]{ \includegraphics[width=0.5\textwidth]{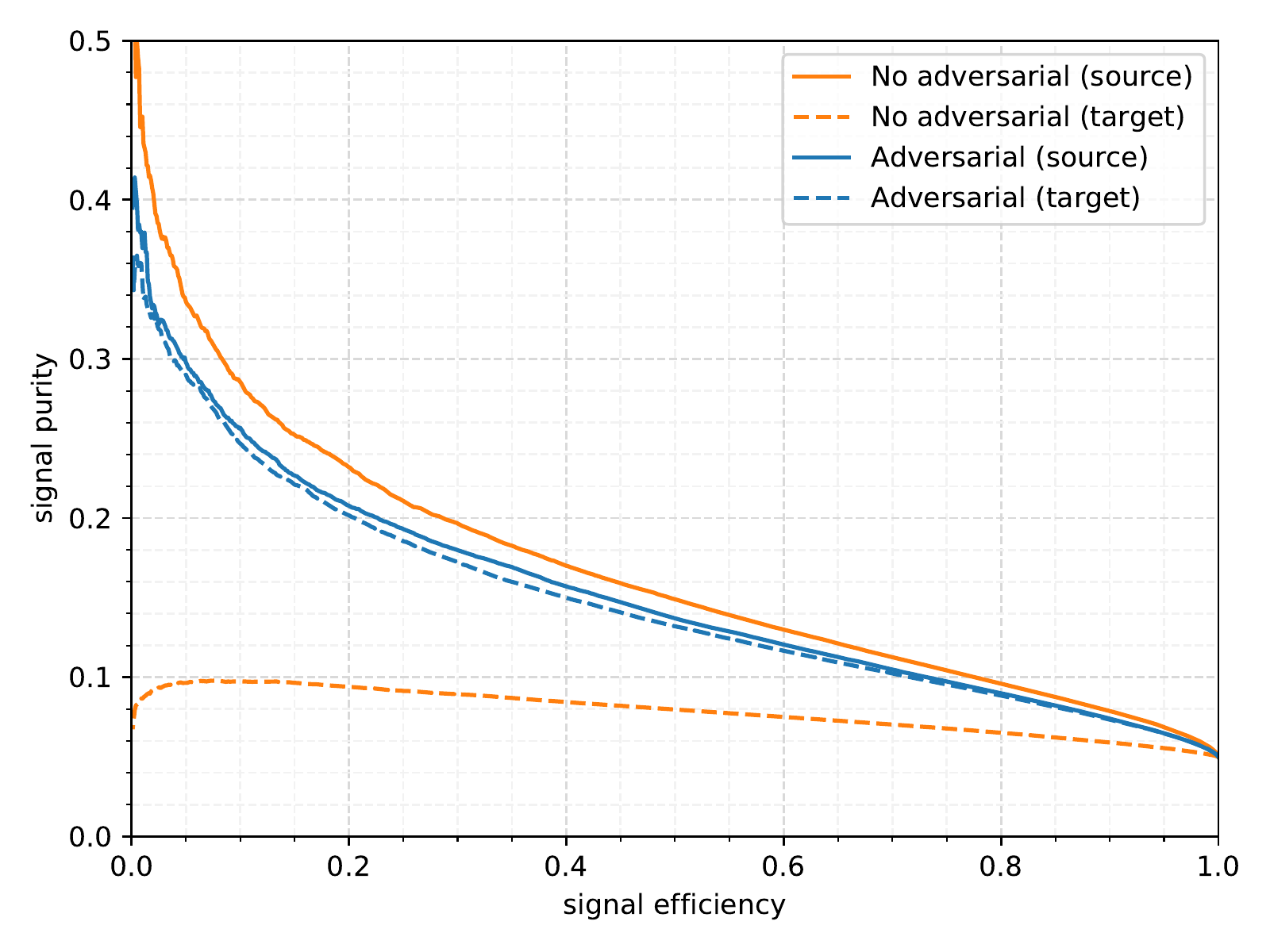}}
	\subfloat[]{\includegraphics[width=0.5\textwidth]{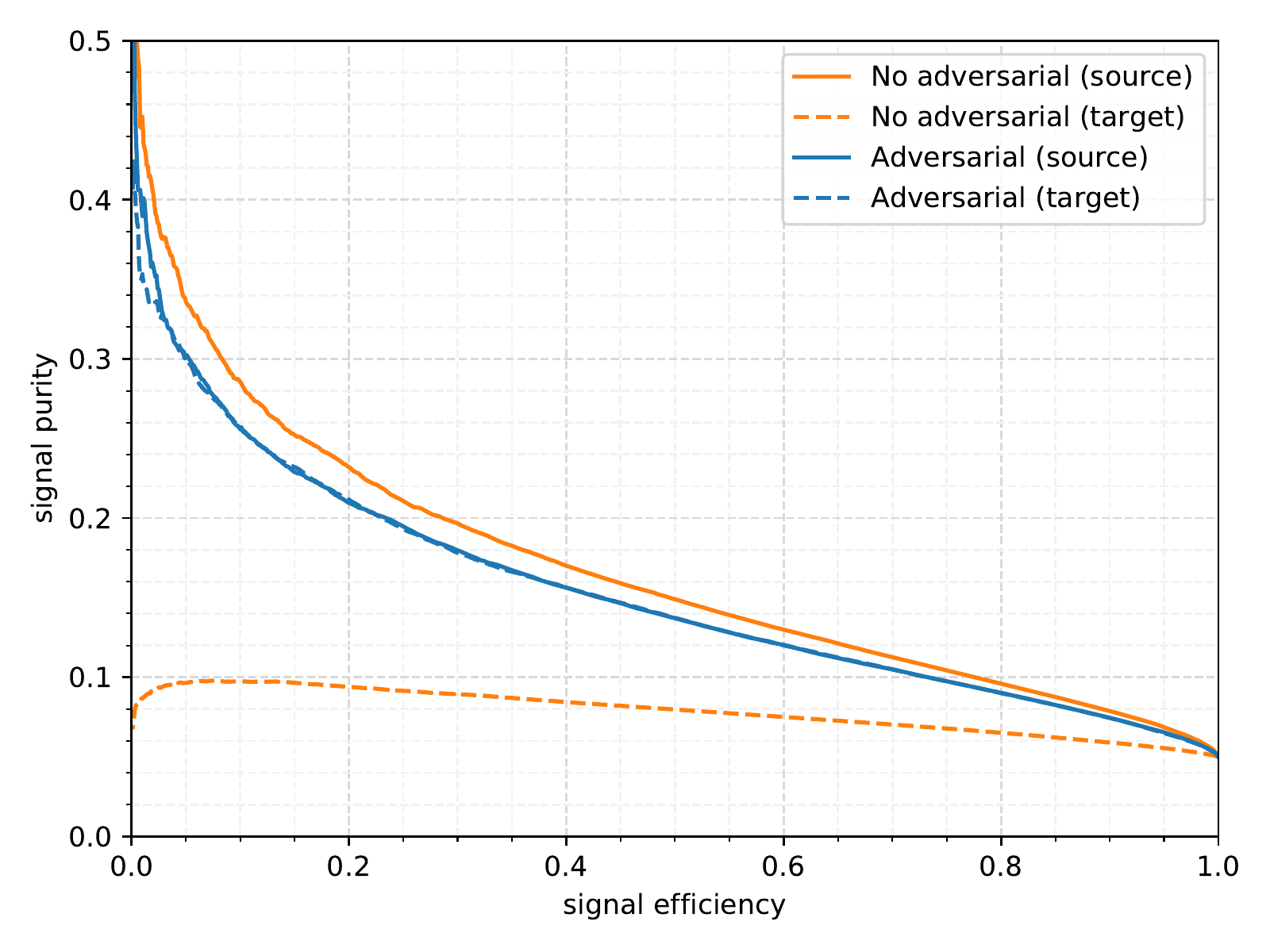}}
	\caption{Signal purity $p_{sig}$ versus signal efficiency $\epsilon_{sig}$ of the label predictor, calculated for the source (solid, upper line) and target (dashed, lower line) sample, with (blue) and without (orange) the adversarial architecture. (a) set-up A (b) set-up B.
	}
	\label{fig:purity}
\end{figure}

Using the optimized $\lambda$ setting we measure the performance in terms of signal purity, which is related to the sensitivity of the measurement. It is defined as the ratio between number of signal events ($s$) and the total of events ($s+b$) that fall above a specific cut in the \emph{label predictor} response: ${p_{sig}=\frac{s}{s+b}}$.
Each possible cut corresponds to a signal efficiency ($\epsilon_{sig} = \frac{s}{N_s}$), which is defined as the fraction of signal selected ($s$) from the total number of signal events ($N_s$). Figure~\ref{fig:purity} shows the whole profile of the signal purity as a function of the signal efficiency. The expected signal to background composition of 5:95 is taken into account.
Classification without the adversarial part reaches around a $9\%$ higher purity on the source domain, but very low values for the target domain. The adversarial network yields very close values for both source and target domains.

To give a numerical example taking the signal purity as an approximation of the analysis sensitivity, we take the results for the source domain as the central value and the difference between the two domains as a 1$\sigma$ uncertainty. For $\epsilon_{sig}=0.5$ we get:
\begin{itemize}
	\item no adversarial network: $p_{sig}=0.148 \pm 0.069$
	\item adversarial set-up A: $p_{sig}$= $0.137 \pm 0.005$
	\item adversarial set-up B: $p_{sig}$= $0.1369 \pm 0.0004$
\end{itemize}
The relative uncertainty  due to the choice of the background model on the signal purity, ignoring other sources of uncertainty, can be improved from 47\% to 4\% (0.3\%) by employing the adversarial network in set-up A (set-up B).

\subsection{Extension of the method towards training with real collision data}
\begin{figure}[!tb]
  \subfloat[]{\includegraphics[width=0.5\textwidth]{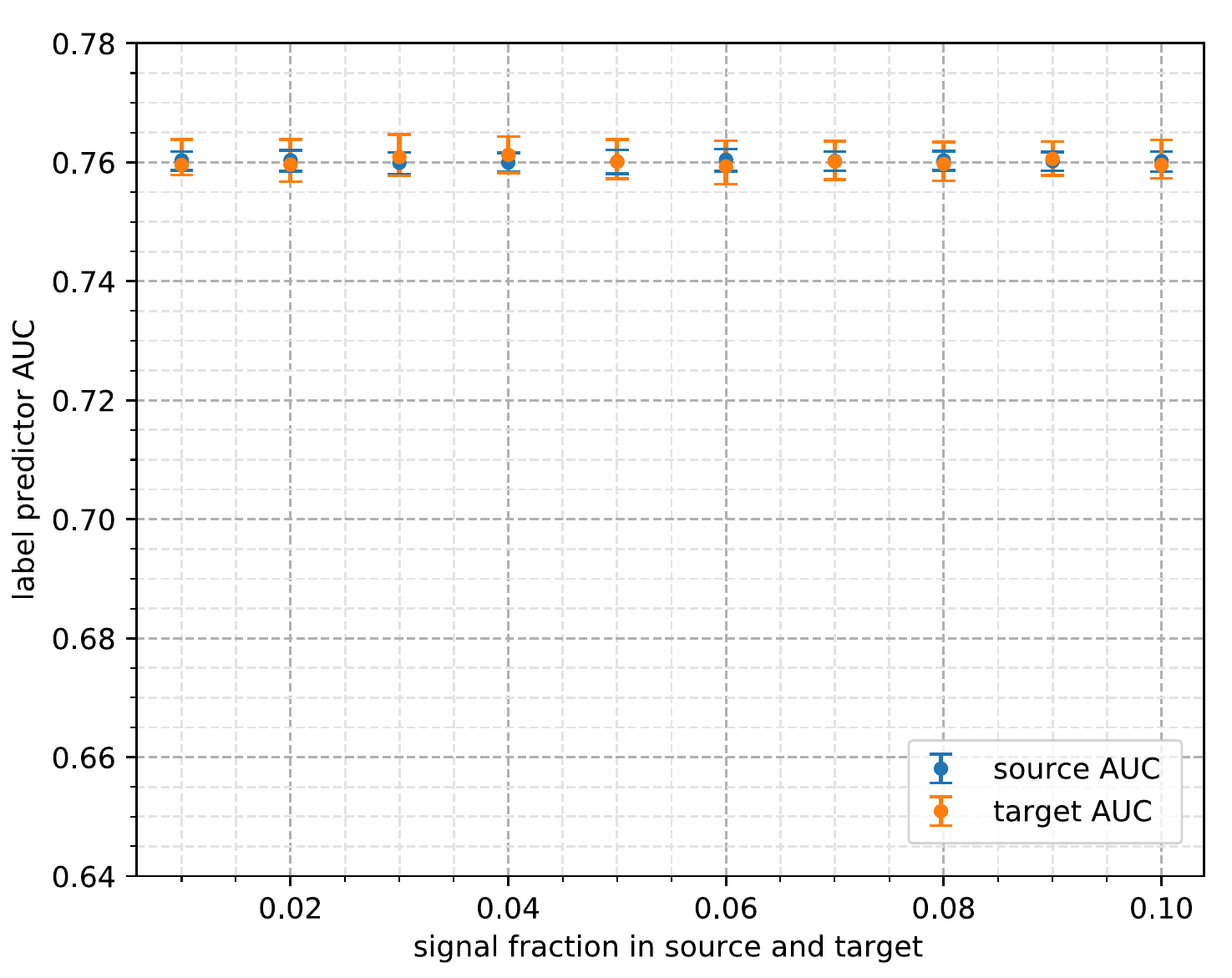} }
  \subfloat[]{\includegraphics[width=0.5\textwidth]{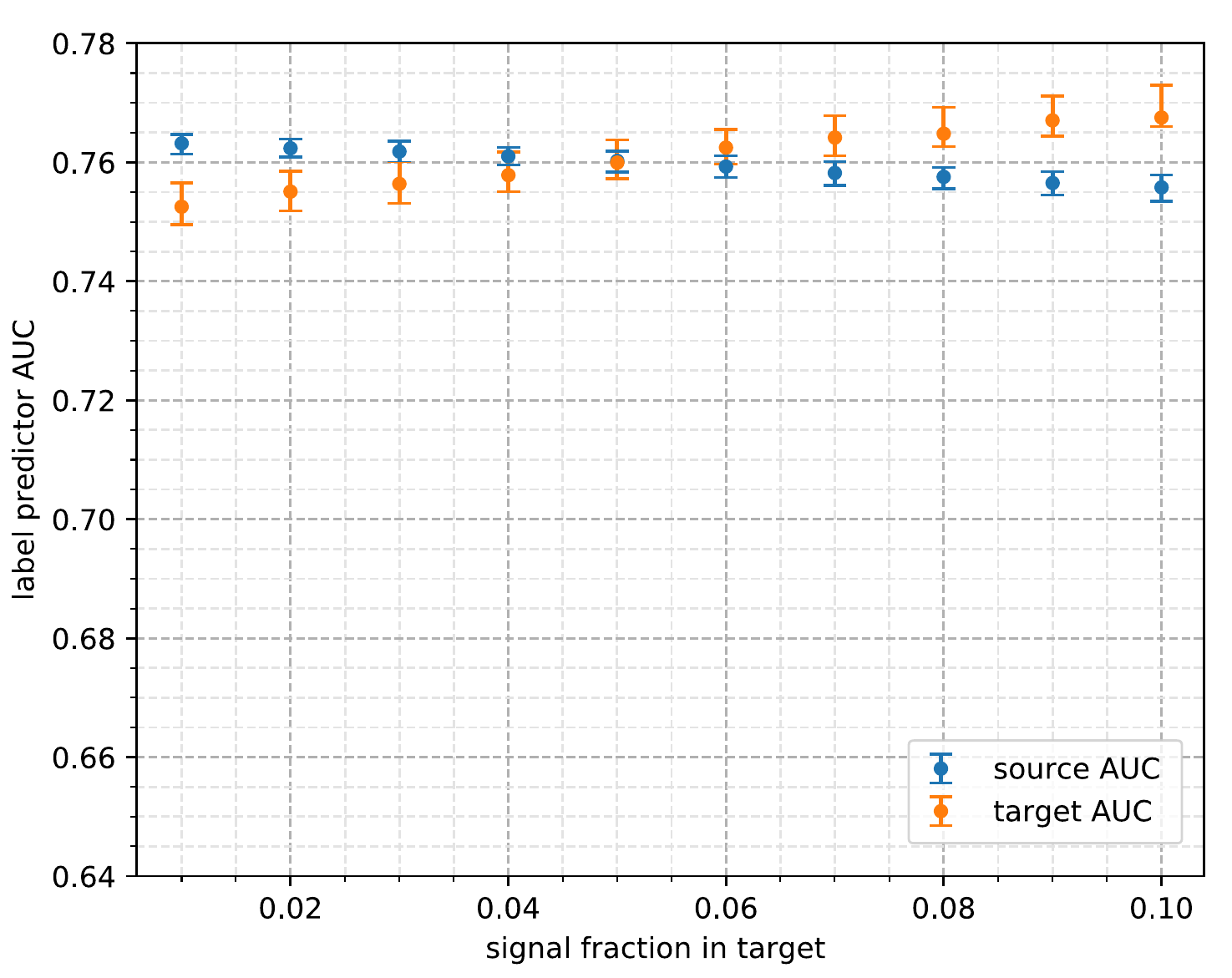} }
  \caption{AUC ROC for the label predictor (a) as a function of the signal fraction in source and target and (b) as a function of the varying signal fraction in target with a fixed 5\% signal fraction in the source. These plots were produced for set-up B with $\lambda=10^{-5}$. Set-up A exhibits a similar behavior.}
\label{fig:signalFraction}
\end{figure}

In this study, no labels were used for computing loss of the domain classifier (except its alignment for signal and background ratio that was so far taken to be the same as for the source domain). One natural extension of the method would be to use real collision data to train domain classifier for adaptation to real data domain. However, in real collision data the ratio of signal to background is only known with limited precision obtained from previous measurements or theoretical predictions. For the results presented so far, the signal to background ratio was set to the predicted value of 5:95 in the target domain, while scaling the source to the same ratio in the \emph{domain classifier}. To check the stability of our results, the  dependence of the \emph{label predictor} output on the chosen signal to background ratio was tested. It was found that a change in its value had no impact, as long as it is the same in both source and target (Fig.~\ref{fig:signalFraction}a).

However, if there is a discrepancy in the signal to background fraction between the two domains, a small bias is introduced. This behavior is shown in Fig.~\ref{fig:signalFraction}b, where a fixed value of 5\% was used for the source signal fraction, while varying signal fraction in the target domain. By varying the signal-to-background ratio by a factor of two away from the ratio in the source domain, a 1.4\% bias was introduced on AUC which is however, still small compared to case without adversarial training. It becomes therefore important to get a good estimate for the signal to background ratio in the target domain when using unlabeled data and to properly account for the effect of this bias on the final result.

We hypothesized that the gap we observe between performance for classifying events in the source or in the target domain when signal and background ratio do not match across domains may be caused by the shift of label distribution. Following number of works attempting to address this issue in unsupervised setting~\cite{conditional}, we applied one such approach to see whether gap issue can be tackled. The implicit alignment approach~\cite{Jiang2020} points out that, among other issues, the differences in label distribution may provide a harmful shortcut to identify the respective domain just on the basis of differences in label frequencies. This may strongly impair domain adaptation by ignoring actual differences in data distributions and thus not handling properly data distribution shift. To circumvent that, authors propose creating re-balanced mini-batches for training domain classifier using pseudo-labels delivered by the label predictor for target inputs, arguing for removing label frequency differences between domains in this way. We saw however that generated pseudo labels have very low reliability, which in turn seems to strongly impair the composition of re-balanced mini batches and do not result in reduction of the classification gap between domains in our case - on the contrary, the gap falls back to the state observed without any domain adaptation. This is not suitable for use case of real collision data adaptation in unsupervised setting, and makes the method in the current form rely on faithful estimate of signal to background ratio in the real collision data as pointed out above.

\section{Conclusion}
\label{sec:conclusion}
We successfully built a feed-forward fully connected adversarial neural network for performing domain adaptation on high energy physics data to enable event classification in the search for the $t\bar{t}H(H\rightarrow b\bar{b})$ process at the LHC.
We demonstrate that adding a \emph{domain classifier} sub-network with a gradient reversal layer helps removing training bias while retaining most of the nominal classification power.
We analyzed the dependence on the hyper-parameters of the network.
We studied the training stability issues that appear due to the  addition of a gradient reversal layer.  We demonstrated that by using linear activation and loss functions, stability and convergence can be significantly improved and  better performance of the network can be achieved.

For the example use case of the ttH(bb) analysis, we demonstrate that the adversarial domain adaptation can produce a label predictor that is almost completely  independent of the domain background model while preserving most of the classification power for target domain. We report the improvements using different measures. Taking the expected signal purity for a signal efficiency of 50\% as a proxy measure for the sensitivity of the analysis, the uncertainty due to the choice of background model can be strongly reduced from a 47\% to a 0.3\% with the Monte Carlo samples used in this study. Significant improvements are also reached in the approximated median significance. Although not demonstrated, we do not expect limitations when extending this approach to adapt to multiple alternative domains, i.e. sources of uncertainty, during training. 

Application of our approach to target samples from real collision data was discussed where no explicit label information from target is required for training of the \emph{domain classifier}. For the selection of optimal value for hyperparameter $\lambda$ that controls the impact of adversarial domain classifier on label predictor, we designed a procedure that does not require labeled target data. However, while per input event example labels from the real target are not necessary for training procedure, we show that in absence of a faithful estimate of the signal to background ratio for the real data target domain, misalignment of the signal to background ratio between source and target domains may lead to a small bias in the classification. This small bias and its impact has then to be addressed in a further downstream analysis.

Using a different ratio of signal to background in source and target domains introduces label distribution shift to the original formulation of the problem, in addition to the already existing data distribution shift given by the different background models in the two domains. Handling both data and label distribution shift for domain adaptation is still a largely unresolved problem in machine learning. For the unsupervised domain adaptation setting we have worked with here, our observations with a recently introduced implicit alignment approach~\cite{Jiang2020} that makes use of pseudo-labels suggests that quality of pseudo labels required for such an approach to cope with both data and label distribution shift is not sufficient for our case. Application of our method to real experimental collision data adaptation in unsupervised setting in its current form will have to therefore rely on fair estimates of signal to background ratio in the real data. 

As discussed above, we see the differences in label frequency between the domains which provides a shortcut for domain identification~\cite{Jiang2020} and harms domain adaptation, as one central issue hampering successful domain adaptation given unknown, different signal background ratios in our case. One potential solution that we envisage for the follow-up work may employ a network architecture that uses yet another adversarial branch dealing explicitly with the task to erase harmful signal-background label information from domain classifier. Given the current progress, we anticipate that this and other advanced approaches~\cite{Prabhu2020} will render our method also capable of handling label shift as well and enable successful adaptation to real collision data in fully unsupervised manner.

\acknowledgments
We acknowledge that  Ilyas Fatkhullin contributed at an early stage of the analysis \cite{Ilyas}.
\\
\\

\noindent \textbf{Data availability}\\
The data that support the findings of this study are available upon request from the authors.


\begin{thebibliography}{99}

\bibitem{ttHbb}
M.~Aaboud {\it et al.} [ATLAS Collaboration],
\textit{Search for the standard model Higgs boson produced in association with top quarks and decaying into a $b\bar{b}$ pair in $pp$ collisions at $\sqrt{s}$ = 13  TeV with the ATLAS detector},
Phys.\ Rev.\ D {\bf 97} (2018) no.7,  072016
[arXiv:1712.08895[hep-ex]].

\bibitem{ganin}
Y. Ganin {\it et al.},
\textit{Domain-Adversarial Training of Neural Networks},
Journal of Machine Learning Research 2016, vol. 17, p. 1-35
[arXiv:1505.07818 [stat.ML]] (2015).

\bibitem{bendavid}
Ben-David, Shai and John Blitzer and Crammer, Koby and Fernando Pereira,
\textit{Analysis of Representations for Domain Adaptation},
Advances in Neural Information Processing Systems 19, p137--144 (2007).

\bibitem{englert}
Englert, Christoph and Galler, Peter and Harris, Philip and Spannowsky, Michael,
\textit{Machine learning uncertainties with adversarial neural networks}
Eur. Phys. J. C79,4 (2019).

\bibitem{JetTag}
C. Shimmin {\it et al.},
\textit{Decorrelated Jet Substructure Tagging using Adversarial Neural Networks},
Phys. Rev. D 96, 074034 (2017),
[arxiv:1703.03507 [hep-ex]].

\bibitem{Sirunyan:2019nfw}
A.~M.~Sirunyan \textit{et al.} [CMS],
\textit{``A deep neural network to search for new long-lived particles decaying to jets,''}
Mach. Learn. Sci. Tech. \textbf{1} (2020), 035012
[arXiv:1912.12238 [hep-ex]].



\bibitem{conditional}
\textit{Conditional Adversarial Domain Adaptation}
Long, Mingsheng and CAO, ZHANGJIE and Wang, Jianmin and Jordan, Michael I
in "Advances in Neural Information Processing Systems 31"
eds. S. Bengio and H. Wallach and H. Larochelle and K. Grauman and N. Cesa-Bianchi and R. Garnett.
by Curran Associates, Inc/
p1640ff (2018)
[http://papers.nips.cc/paper/7436-conditional-adversarial-domain-adaptation.pdf]


\bibitem{Pivot}
G. Louppe, M. Kagan and K. Cranmer,
\textit{Learning to Pivot with Adversarial Networks},
Advances in Neural Information Processing Systems 30, p981 (2017),
[arxiv:1611.01046 [stat.ML]] (2016).

\bibitem{Ranking}
P. Glaysher, J. M. Katzy and S. An,
\textit{Iterative subtraction method for Feature Ranking},
[arXiv:1906.05718 [physics.data-an] ](2019).


\bibitem{opendata}
S.~V.~Chekanov [HepSim Group],
Advances in High Energy Physics, vol. 2015, Article ID 136093, 7,
[arxiv:1403.1886 [hep-ph]].

\bibitem{Alwall:2011uj}
J.~Alwall, M.~Herquet, F.~Maltoni, O.~Mattelaer and T.~Stelzer,
JHEP {\bf 1106} (2011) 128,
doi:10.1007/JHEP06(2011)128,
[arXiv:1106.0522 [hep-ph]].

\bibitem{Corcella_2001}
G.~Corcella, I.~Knowles, G.~Marchesini, S.~Moretti, K.~Odagiri, P.~Richardson, M.~Seymour, B.~Webber
JHEP {\bf 0101:}010 (2001),
[arXiv:hep-ph/0011363].

\bibitem{Sjostrand:2006za}
T.~Sjostrand, S.~Mrenna and P.~Z.~Skands,
JHEP {\bf 0605} (2006) 026
[hep-ph/0603175].

\bibitem{Tomas}
T. Jezo, J. Lindert, N. Moretti, S. Pozzorini,
\textit{New NLOPS predictions for $t\bar{t} +b$-jet production at the LHC},
 Eur. Phys, J. C78-6-502 (2018),
 [arXiv:1802.00426 [hep-ph]].

\bibitem{delphes}
J.~de Favereau {\it et al.} [DELPHES 3 Collaboration],
\textit{DELPHES 3, A modular framework for fast simulation of a generic collider experiment},
JHEP {\bf 1402} (2014) 057,
[arXiv:1307.6346 [hep-ex]].

\bibitem{antikt}
M.~Cacciari, G.~P.~Salam and G.~Soyez,
JHEP \textbf{04} (2008), 063
[arXiv:0802.1189 [hep-ph]].

\bibitem{Aaboud:2018xwy}
M.~Aaboud {\it et al.} [ATLAS Collaboration],
\textit{Measurements of b-jet tagging efficiency with the ATLAS detector using $ t\overline{t} $ events at $ \sqrt{s}=13 $ TeV},
JHEP {\bf 1808} (2018) 089
[arXiv:1805.01845 [hep-ex]].

\bibitem{keras}
F. Chollet {\it et al.},
\textit{Keras},
https://github.com/fchollet/keras (2015).

\bibitem{tensorflow}
M.~Abadi et.al.,
\textit{TensorFlow: Large-Scale Machine Learning on Heterogeneous Systems},
[https://www.tensorflow.org/]

\bibitem{xavier}
X. Glorot, Y. Bengio,
\textit{Understanding the difficulty of training deep feedforward neural networks},
Proc. of 13th Intern.Conf.on Artificial Intelligence and Stat.(AISTATS) Vol.9 of JMLR (2010).

\bibitem{hyperopt}
J. Bergstra, D. Yamis, D.D. Cox,
\textit{Making a Science of Model Search},
In Proc. of 30th Int. Conf. on Machine Learning (ICML 2013)
 Volume 28 p.I-115 (2013).

\bibitem{adam}
Diederik P. Kingma, Jimmy Ba,
\textit{Adam a method for stochastic optimisation}
\arxivnumber{1412.6980}

\bibitem{cowan}
C.Adam-Bourdarios, G.Cowan, C. Germain, I.Guuon, B. Kegl, D.Rousseau,
JMLR: Workshop and Conference Proceedings: 42:19-55 (2015)

\bibitem{Ilyas}
I. Fatkhullin,
DESY Summer Student Program's report (2019)
[https://www.desy.de/f/students/2019/reports/Ilyas.Fatkhullin.pdf]

\bibitem{Jiang2020}
X. Jiang, Q. Lao, S. Matwin, M. Havaei, 
\textit{Implicit class-conditioned domain alignment for unsupervised domain adaptation} 
International Conference on Machine Learning (PMLR), 2020, 4816-4827 

\bibitem{Prabhu2020}
V. Prabhu, S. Khare, D. Kartik, J. Hoffman, 
\textit{SENTRY: Selective Entropy Optimization via Committee Consistency for Unsupervised Domain Adaptation}, [arXiv:2012.11460 [cs.CV]], 2020



\end{thebibliography}
\end{document}